\documentclass[letterpaper, 10 pt, conference]{ieeeconf}  

\IEEEoverridecommandlockouts                              

\usepackage{graphicx}
\usepackage{multirow}
\usepackage{booktabs}
\usepackage{amsmath}
\usepackage{mathtools}
\usepackage{fourier}
\usepackage{stfloats}
\usepackage{caption}

\newcommand{\commentOUT}[1]{}

\usepackage[dvipsnames,svgnames,x11names]{xcolor}
\usepackage[markup=underlined]{changes}
\definechangesauthor[color=BrickRed]{JK}
\definechangesauthor[color=NavyBlue]{NN}
\usepackage{todonotes}
\setcommentmarkup{\todo[color={authorcolor!20},size=\scriptsize]{#3: #1}}

\title{\LARGE \bf Structured Spatial Reasoning with Open Vocabulary Object Detectors}

\author{Negar Nejatishahidin, 
Madhukar Reddy Vongala, Jana Kosecka
 \\
 George Mason University, USA \\
\texttt{\{nnejatis, mvongala, kosecka\}@gmu.edu} 
}

\begin{document}
\maketitle 
\thispagestyle{empty}
\pagestyle{empty}

\begin{abstract}
Reasoning about spatial relationships between objects is essential for many real-world robotic tasks, such as fetch-and-delivery, object rearrangement, and object search. The ability to detect and disambiguate different objects and identify their location is key to successful completion of these tasks. Several recent works have used powerful Vision and Language Models (VLMs) to unlock this capability in robotic agents. In this paper we introduce a structured probabilistic approach that integrates rich 3D geometric features with state-of-the-art open-vocabulary object detectors to enhance spatial reasoning for robotic perception. The approach is evaluated and compared against zero-shot performance of the state-of-the-art Vision and Language Models (VLMs) on spatial reasoning tasks. To enable this comparison, we annotate spatial clauses in real-world RGB-D Active Vision Dataset~\cite{DBLP:journals/corr/AmmiratoPPKB17} and conduct experiments on this and the synthetic Semantic Abstraction~\cite{ha2022semantic} dataset. Results demonstrate the effectiveness of the proposed method, showing superior performance of grounding spatial relations over state of the art open-source VLMs by more than 20\%.

\end{abstract}


\section{INTRODUCTION}

\begin{figure}[t]
\centering
\includegraphics[width=\linewidth]{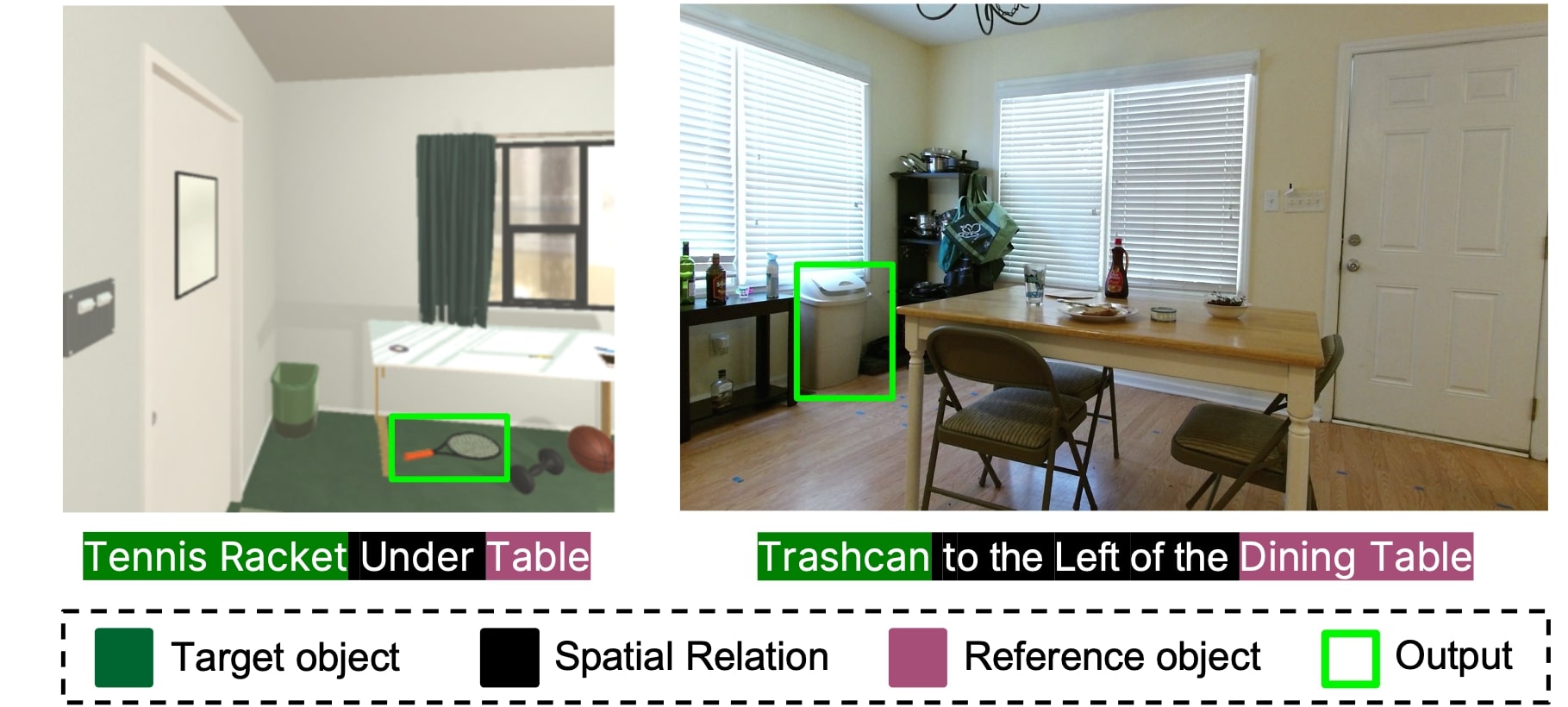}
\caption{Examples showing inputs and desired outputs. The input consists of a triplet <target object, spatial relation, reference object> and corresponding image. The output is the target object bounding box. }
\label{fig:problem_def}
\end{figure}

The task of spatial relationship detection refers to the ability to localize objects of interest and determine the spatial relationships between them. This capability is essential for interpreting the physical world and understanding environment layouts. 
Examples of spatial clauses {\em 'tennis racket under the table'} or {\em 'trashcan left of the dinning table'} are in Figure \ref{fig:problem_def}. In robotic settings the need to recognize and ground the spatial clauses is encountered in Visual Question Answering (VQA)~\cite{Chen2024SpatialVLMEV}, human-robot interaction, task planning \cite{mo2022towards}, object rearrangement, fetch-and-delivery~\cite{Yuan2024RoboPointAV}, object search, and scene understanding \cite{Jiang2023HierarchicalRA, nejatishahidin2023graph} tasks.
%

Rapid advancements in Vision and Language Models (VLMs) that are trained in self-supervised setting using large amounts of image-caption pairs~\cite{Lu2019ViLBERTPT, DBLP:journals/corr/abs-2112-03857, Tan2019LXMERTLC, Li2021AlignBF, liu2024improved, liu2024visual, team2023gemini, anil2023palm, radford2021learning, alayrac2022flamingo} marked notable improvements in large body of previously challenging vision and language tasks. For these new approaches, reasoning about spatial relationships in a zero-shot manner or by fine-tuning the large vision and language models (VLMs) continues to be a challenging task~\cite{kamath-etal-2023-whats, Rajabi2023TowardsGV, Liu2022VisualSR}. 
The limited spatial reasoning capability of VLMs was noted in SpatialVLM~\cite{Chen_2024_CVPR}. The authors pointed out the lack of training examples with 3D spatial knowledge and proposed fine-tuning PALM-E~\cite{Driess2023PaLMEAE}, an embodied multi-modal model, with a large dataset of image-text pairs containing descriptions of spatial relations. 
While the approach is effective, fine-tuning these models is resource-intensive and demands significant data engineering.  
This motivates us to explore traditional structured methods in combination with the state-of-the-art modules and explore the advantages over foundational models in controlled settings. In this work we propose a modular approach that leverages rich 3D geometric features to enhance spatial relationship detection, introduce a new dataset and compare the performance with traditional VLM models. The proposed approach for data labeling can be viewed both as a standalone model or automated method for generating training data to fine-tune VLMs for specific environments, similar to SpatialVLM~\cite{Chen_2024_CVPR}.   

Previous approaches for spatial relations learning and evaluation~\cite{Johnson2016CLEVRAD, Yang2019SpatialSenseAA, Goyal2020Rel3DAM} used images from several computer vision benchmarks. In~\cite{ha2022semantic}, authors introduced synthetic spatial relation dataset for robotics using AI2-THOR~\cite{Kolve2017AI2THORAI} simulator. There are currently no real-world RGB-D datasets available for spatial reasoning. We extended the RGB-D Active Vision Dataset (AVD) \cite{DBLP:journals/corr/AmmiratoPPKB17} to include spatial relations annotations. AVD is a public RGB-D dataset for active robotic vision tasks, consisting of dense scans of real indoor environments. We have proposed a fully automated approach for annotating spatial relations, that can also be applied to any specific environment with high-performing object detectors. 
%
Our main contributions are: 
\begin{itemize}

\item Structured probabilistic approach utilizing the state-of-the-art open-set vocabulary object detector \cite{zhou2022detecting, Liu2023GroundingDM, DBLP:journals/corr/abs-2112-03857} and relying on 3D geometric cues to recognize spatial relations.
\item Novel geometric Spatial Relation Classification Model that takes as input the pose and dimensions of objects computed from their 3D point clouds.
\item Probabilistic Ranking Module that combines the evidence from object grounding and spatial relationship classification to get the best triplet.
\item Technique for automated labeling of the AVD dataset for spatial reasoning containing more than 1,500 images and 50,000 expressions.
\end{itemize}
We demonstrate the results on both the real AVD-Spatial dataset and the synthetic Semantic Abstraction dataset, showcasing the effectiveness of our approach and its superior performance over GroundingDINO~\cite{Liu2023GroundingDM} and LLaVA \cite{liu2024visual} models. Additionally, we conducted a thorough ablation study to evaluate the effectiveness of 2D, 3D features and language priors.
 
\begin{figure}[t]
  \centering
    \includegraphics[width=\linewidth]{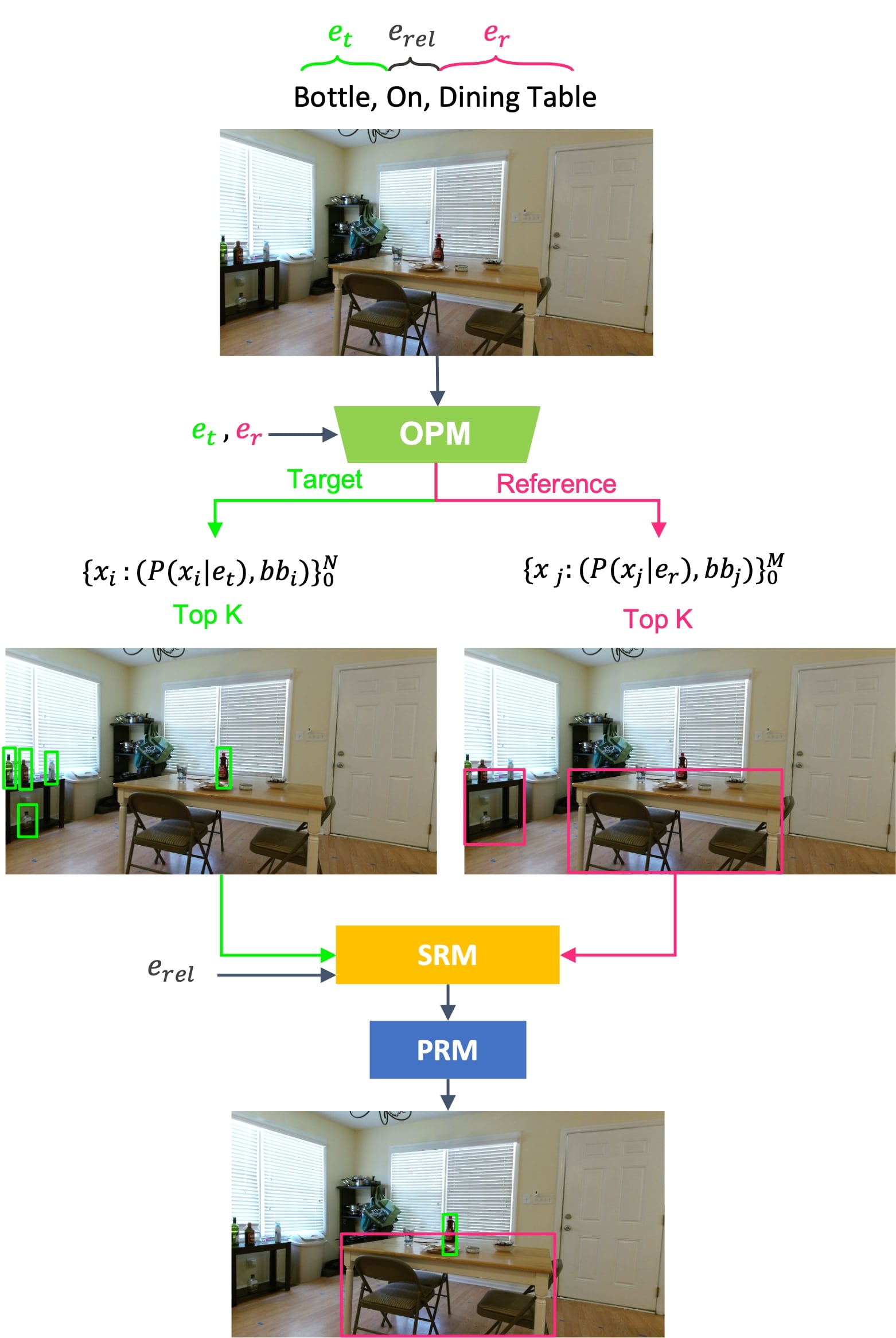}
    \caption{This is an overview of our pipeline. It consists of three main modules: first, the Object Proposal Module (OPM), which provides a set of boxes as candidates for the target and reference objects; second, the Spatial Relation Module (SRM), which outputs a distribution over possible relationships for each pair; and third, the Probabilistic Ranking Module (PRM), which identifies the best triplet.}
   \label{fig:approach}
\end{figure}

\section{Related work}

Two lines of related work to spatial relationship recognition are {\em structured semi-supervised learning} approaches, where spatial relationships are predicted compositionaly in a semi- supervised setting and {\em vision-language models}, that rest on capabilities of large foundational models pre-trained on large amount of image-text pairs. 

More recent structured approaches to spatial reasoning separate the problems of object detection and determining the spatial relationships between objects. Rel3D~\cite{Goyal2020Rel3DAM} demonstrated the effectiveness of 3D geometric features in simulation for spatial relation recognition. In~\cite{mo2022towards}, the authors propose a probabilistic modular technique for visual question answering (VQA) using CLIP~\cite{clip} for object detection in tabletop robotic manipulation tasks. The visual setting of this approach is quite simple, involving unoccluded objects and considering only 2D information. The approaches of~\cite{ha2022semantic, Subramanian2022ReCLIPAS}, show greater effectiveness in spatial reasoning either using additional training or by decomposing the problem structure, incorporating state-of-the-art models like CLIP~\cite{clip} and GLIP~\cite{ DBLP:journals/corr/abs-2112-03857}, for detecting objects. 
They naturally outperform zero-shot performance of object detectors that treat the spatial phrases as referring expressions. 
For instance, in~\cite{ha2022semantic}, a module was proposed for obscured object localization using CLIP to compute relevancy maps from RGB-D data, followed by training a spatial reasoning module. Although performance on spatial relationship recognition remained low (<30\%), it outperformed CLIP’s zero-shot capability.
In the proposed work we leverage state-of-the-art object detectors~\cite{zhou2022detecting, Liu2023GroundingDM} and introduce a novel spatial relation module using 3D geometric features. 

With the advent of large vision-language models, the main downstream tasks, including image captioning, image classification, image-text matching, and visual question answering (VQA), have seen notable improvement. In~\cite{Liu2022VisualSR}, the authors introduced a new Visual Spatial Reasoning (VSR) dataset for evaluating spatial reasoning capabilities and used image-text matching to assess the performance of VLMs, which proved to be ineffective even after fine-tuning. In~\cite{Rajabi2023TowardsGV} the authors compared the performance of several other VLMs~\cite{Tan2019LXMERTLC, gpv, mdetr} on the VSR dataset and presented fine-grained compositional grounding of spatial relationships with a ranking of spatial clauses. \\
The most recent representative of open-source VLMs is LLaVA, LLaVAnext \cite{liu2024visual}. LLaVA uses the CLIP visual encoder (ViT-L/14) for encoding images and the Vicuna language model for text generation, linked by a trainable projection layer that ensures seamless multi-modal interaction. A key feature is LLaVA’s innovative data curation, leveraging GPT-4 to generate high-quality instruction-following data, including conversational queries and complex reasoning tasks, closely mimicking real-world interactions. SpatialVLM \cite{Chen_2024_CVPR} attributes low performance on multi-modal models in spatial reasoning tasks to the lack of adequate spatial relation training data. The approach achieves notable improvements by fine-tuning PALM-e foundational model \cite{Driess2023PaLMEAE} with image-text pairs containing spatial relations. However, training similar models requires a significant amount of data labeling, engineering, and computational power. Fine-tuning VLMs for downstream tasks also often results in using overly large and complex models for specific tasks.\\ 
\noindent 
To evaluate the proposed approach, several datasets are available, each with its own strengths and limitations. Semantic Abstraction~\cite{ha2022semantic} and Rel3D~\cite{Goyal2020Rel3DAM} are synthetic datasets with fewer than 10 spatial relations. While they provide ground truth depth images and bounding box annotations, their synthetic nature avoids the challenges of real-world environments, such as noisy measurements and large amount of clutter. SpatialSense dataset \cite{Yang2019SpatialSenseAA} has real images and bounding box annotations, but lacks depth data and is not targeted towards robotic applications. Visual Spatial Recognition (VSR) dataset (\cite{Liu2022VisualSR} has real images with more advanced expressions, but lacks bounding box annotations. Visual Genome \cite{DBLP:journals/corr/KrishnaZGJHKCKL16} provides real data and annotations, but it lacks 3D information and exhibits imbalance in spatial relations. Importantly, none of these datasets are situated in a robotic setting where interaction with the scene and traversal in the environment is feasible. Therefore, we proposed an Active Vision Dataset Spatial (AVD-Spatial) benchmark \cite{DBLP:journals/corr/AmmiratoPPKB17}, addressing the limitations present in existing datasets and facilitating research on spatial reasoning in robotic settings.

\section{APPROACH}
\label{sec:approach}
The proposed approach comprises three core components: the object proposal module (OPM), which utilizes an open-vocabulary object detector for target and reference objects detection; the spatial relationship module (SRM) that classifies spatial relationships between pairs of objects in the scene; and the probabilistic ranking module (PRM) for final target object ranking. The approach is visualized in Figure~\ref{fig:approach}.
%
Some examples from the datasets are shown in Figure \ref{fig:problem_def}. The training set consists of an expression \( e \) and its corresponding image \( I \). The expression \( e = (e_r, e_{rel}, e_t)  \) in natural language consists of three components: a reference object \( e_r \) and target object \( e_t \)  and their spatial relation \( e_{rel} \). The goal is to determine the precise location of the referred object described in the spatial expression, e.g. {\em tennis racket under the table} in the image. This task is closely related to recent formulations demonstrated in previous works \cite{ha2022semantic, Rajabi2023TowardsGV, Yang2019SpatialSenseAA, Liu2022VisualSR, Goyal2020Rel3DAM}.\\
\noindent
\textbf{Object Proposal Module.} 
To detect a referred and target object we use open set vocabulary object detector \( f_{d} \). We have used only the target and reference object class names as the vocabulary set and run it separately. For each target and reference object \( f_{d} \) outputs set of $N$ candidate objects called ${x_i}_1^N$ containing the bounding box $\{bb_i\}_1^N$ with corresponding confidence scores \(\{P(x_i|e_t)\}_1^N\) or \(\{P(x_i|e_r)\}_1^N\) for detected objects. We sorted the predictions based on the confidence scores. The top $K$ most probable objects will be considered as the target object candidates for the subsequent experiments, as shown in Figure \ref{fig:approach}. 

\begin{figure}[t]
  \centering
    \includegraphics[width=\linewidth]{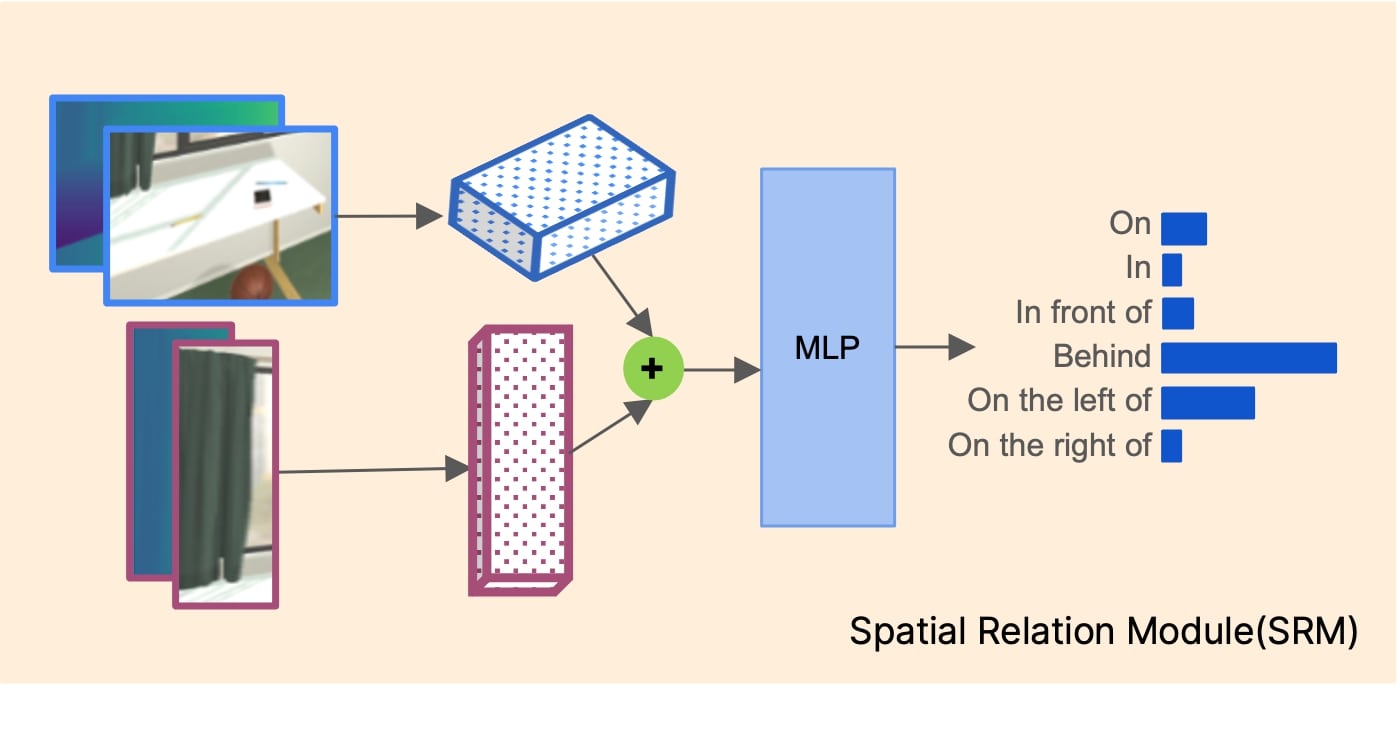}
    \caption{The image and depth data are masked with the object mask to compute the 3D point cloud. PCA fits a box to the point cloud, and the 6D pose and bounding box dimensions of both objects are concatenated as inputs to the MLP, which outputs a distribution over spatial relation classes.}
   \label{fig:SRM}
\end{figure}

\noindent
\textbf{Spatial Relation Module.} 
For spatial relationship classification, we train a multi-layer perceptron MLP to estimate probabilities of different spatial relationships. For a pair of bounding boxes  $\phi(bb_i)$ and  $\phi(bb_j)$, one target and one reference object, we first compute 3D geometric features for each box $\phi(bb_i)$. 
We use camera intrinsic parameters and bounding box's segmentation mask\footnote{In the case of GroundingDINO as the object detector, we used SAM \cite{Kirillov2023SegmentA} to get the segmentation mask for every box} and depth map to reconstruct the 3D point cloud of the object. By applying PCA on the 3D point cloud, we estimated the orientation and translation of the object's coordinate frame, as well as the dimensions of the 3D bounding box with $bb  = \left\{ T = (x, y, z), R \in SO(3),  D = (\ell, w, h) \right\}$ where: \(T\) is the translation vector, \(R\) is the rotation matrix, \(D\) denotes the dimensions. For each pair $bb_i$ and $bb_j$ we flatten and concatenate them to form the 2D/3D input features to MLP.  
The model has been trained to classify these features into 6 classes for Semantic Abstraction, 'on, in, to the left of, to the right of, behind, in front of' and 'above, below, to the left of, to the right of, behind, in front of'  for AVD-Spatial, Figure \ref{fig:SRM} visualizes this process.
\noindent
Additional experiments were conducted using only 2D features, to evaluate the role of 3D information. We utilized 2D box centers and box dimensions $ bb = (x_c, y_c, w, h) $ to demonstrate the effectiveness of model. To understand the impact of linguistic information, on MLP performance, object class names encoded using FastText were concatenated to the MLP input alongside the geometric features.
This lead to model with language priors. 
The results of these variations of SRM module on Semantic Abstraction and AVD-Spatial Datasets are presented in Tables~\ref{table:cls_detic} and \ref{table:MLP_AVD}.

\noindent
\textbf{Probabilistic Ranking Module.} Object proposal module identifies the top $K$ candidate bounding boxes for target and reference objects and SRM predicts probabilities between each pair of bounding boxes $ P(\phi(box_i), \phi(box_j)|e_{rel})$.
To rank the best pair as target and reference objects, assuming the independence of $e_t, e_r, e_{rel}$ the probability of the 
object pair given the entire referring expression is computed as
\begin{equation}
    \begin{array}{cc}
    P(x_i, x_j|e ) & \propto P(x_i|e_t)P(x_j|e_r)P(x_i,x_j|e_{rel})\\
    \end{array}
    \label{eq:1}
\end{equation}
\noindent
Where $x_i$ is a candidate for the target object and $x_j$ is a candidate for the reference object, the relation probabilities between each pair of boxes using the MLP and obtain the probability $ P(\phi(bb_i), \phi(bb_j)|e_{rel})$.  The $P(x_i|e_t)$ and $P(x_j|e_r)$ are  confidence scores of the object detector. Based on SRM,
\begin{equation}
\begin{array}{cc}
  P(x_i,x_j|e_{rel}) & \propto P(\phi(bb_i), \phi(bb_j)|e_{rel})
\end{array}
\end{equation}

\section{Datasets}
We evaluated our approach on synthetic Semantic Abstraction \cite{ha2022semantic} dataset and  spatial relation detection dataset built on top of AVD~\cite{DBLP:journals/corr/AmmiratoPPKB17}. 
\subsection{Semantic Abstraction}
\begin{figure}[t]
  \centering
    \includegraphics[width=0.9\linewidth]{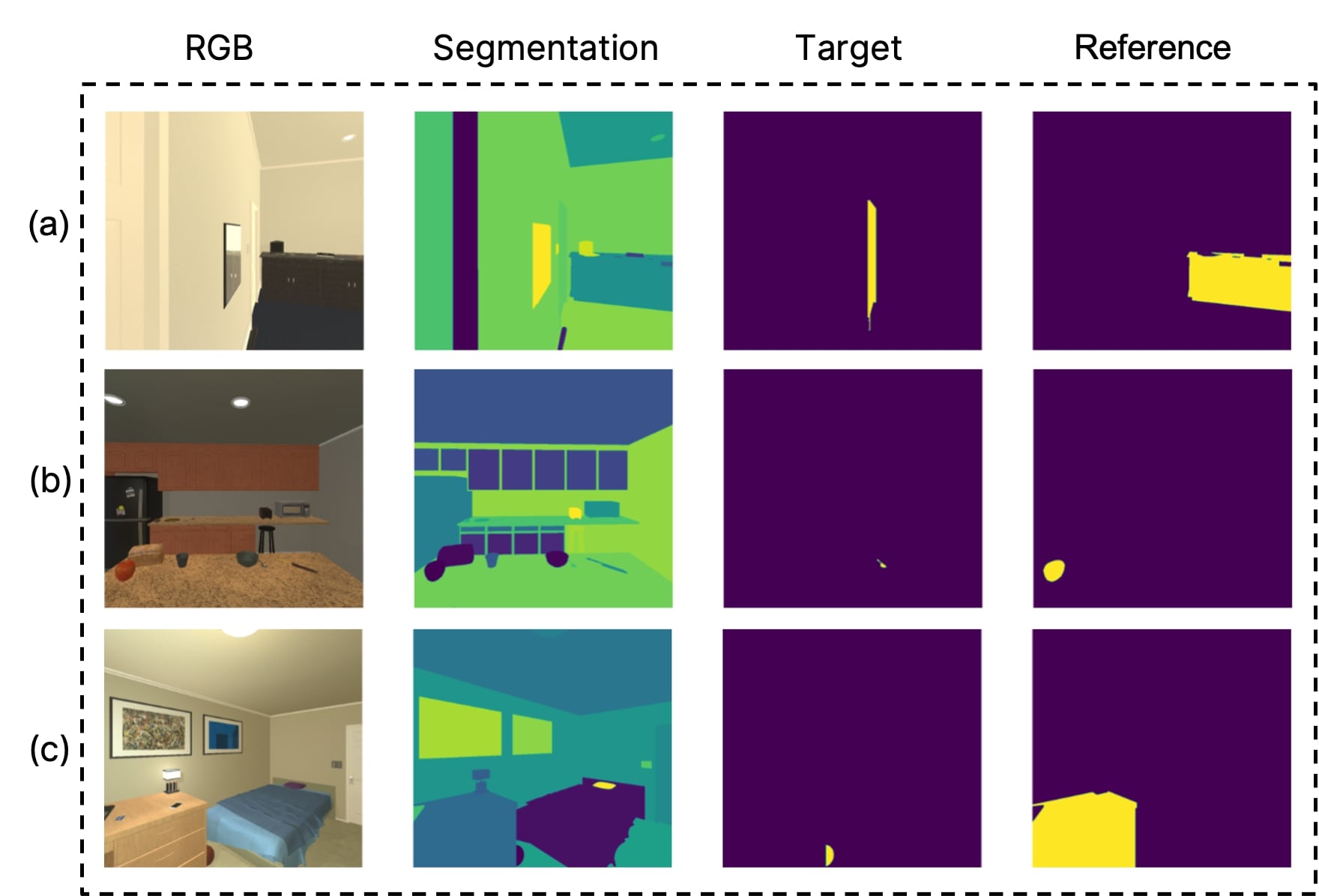}
    \caption{Examples from Semantic Abstraction dataset. (a) Door on the left of the dresser. (b) Spoon on the right of the apple. (c) Basketball on the right of desk. The dataset showcases challenges arising from small object sizes, occlusions, and clutter. Each data sample has semantic segmentation, depth images, ground truth mask of objects, and the expressions.}
   \label{fig:sampleimages}
\end{figure}

This dataset comprises a total of 6085 views spread across 100 scenes. The evaluation was conducted on three main subsets: novel visuals, novel synonyms, and novel classes, consisting of 1244, 940, and 597 views, respectively, distributed across 20 test scenes. The category of novel visuals presents a test set in which the texture and visual appearance of the objects are novel compared to the training set. In the novel synonyms test set, synonyms of object classes are used. In the novel classes set, new object classes are presented. Using these subsets, we can demonstrate the effectiveness of the proposed approach when encountering visual and textual domain gaps. The dataset features six common spatial prepositions: {\em behind, left of, right of, in front of, on top of, inside}.Examples and difficulties of the dataset are highlighted in Figure~\ref{fig:sampleimages}. 

\subsection{Spatial Active Vision Dataset}
\begin{figure*}[t]
  \centering
    \includegraphics[width=0.9\linewidth]{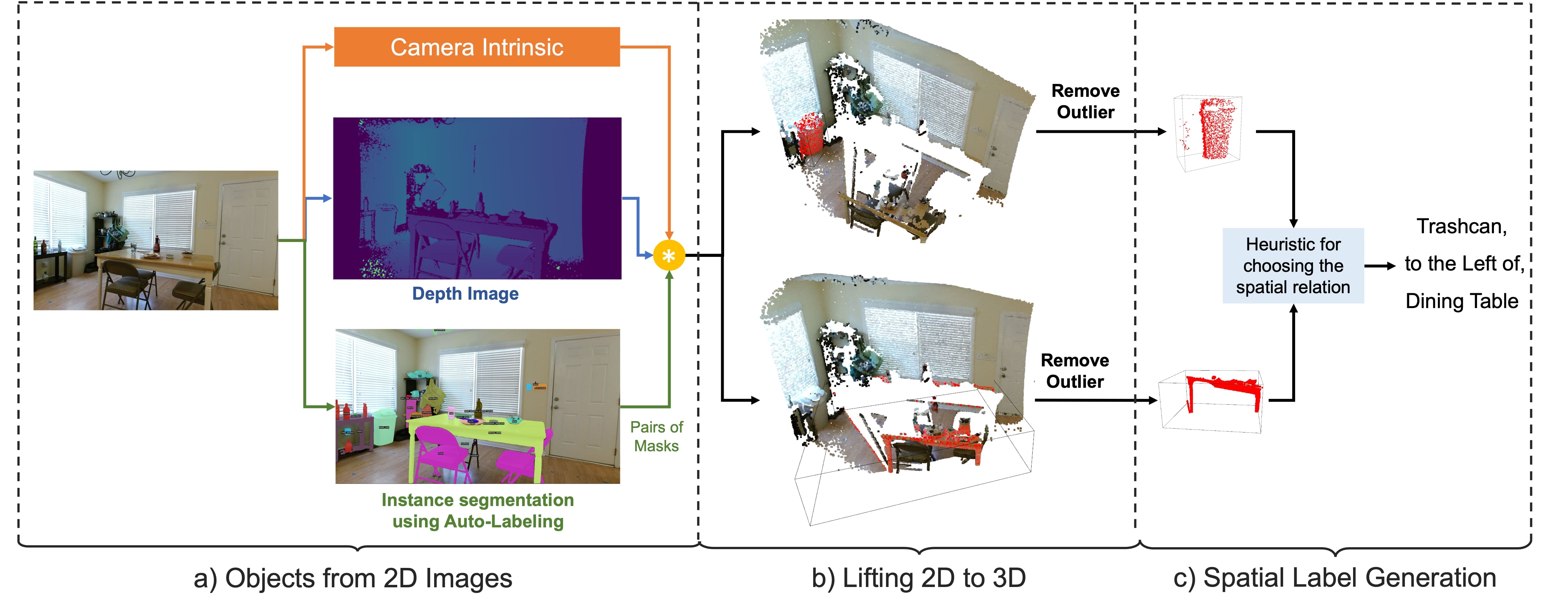}

    \caption{In part (a), we generate instance segmentation using auto-labeling \cite{Li2023LabelingIS} to obtain object pairs. In part (b), we generate object point clouds, remove outliers, and fit 3D bounding boxes. Finally, in part (c), we generate spatial reasoning expressions based on the 3D information.}
   \label{fig:labeling_pipline}
\end{figure*}

The Active Vision Dataset (AVD) \cite{DBLP:journals/corr/AmmiratoPPKB17} consists of dense scans of real indoor environments with available depth images and camera information. The dataset includes a total of 17 scenes, each offering multiple viewpoints of the scene as the robot traverses the environment. In many images objects may be occluded, present in cluttered environments, or appear small due to distance in the image. 
AVD Spatial relation dataset, uses an automated approach for annotating spatial relations in the images inspired by SpatialVLM \cite{Chen2024SpatialVLMEV}. 
Pseudo-labelling of AVD Spatial proceed as follow:
\begin{itemize}
    \item \textbf{2D Objects Detection};
    To extract the segments, bounding boxes, and class labels of the objects in the scene, we employed the auto-labeling pipeline proposed in \cite{Li2023LabelingIS}. This approach leverages DETIC \cite{zhou2022detecting}, the state-of-the-art semantic segmentation model MaskFormer \cite{Cheng2021PerPixelCI}, and the foundational Segment Anything Model (SAM) \cite{Kirillov2023SegmentA} to annotate the image with pixel-wise labels. The pipeline provides both instance and semantic segmentation.
    \item \textbf{Lifting 2D to 3D}; Using ground truth depth and camera intrinsics, we transformed 2D information into object-centric 3D point clouds. We addressed issues with inaccurate segmentation boundaries and missing depth values by cleaning and denoising the point clouds. We fitted 3D bounding boxes to the objects using PCA.
    \item \textbf{Spatial Reasoning Expression Dataset}; To generate qualitative expressions, we select pairs of objects 
    along with their corresponding 3D point clouds and bounding boxes, to create 3D spatial expressions for AVD.
\end{itemize}
\noindent
We annotated 50,000 spatial expressions across six spatial relations—{\em right, left, above, below, behind, in front of} — for the first scene of AVD, which includes 1,500 images and 400 different object classes. The approach is visualized in Figure \ref{fig:labeling_pipline}. We refer to it as AVD-Spatial. 

\section{Experiments}
\label{sec:experiments}


\subsection{Training}
The Spatial Relation Module (SRM) employs a three-layer MLP, with the learning rate set at 0.001 using the ADAM optimizer. A learning rate decay of 0.5 was applied after every three epochs, for a total of 10 epochs across both datasets. For Semantic Abstraction, we utilized multiclass classification with cross-entropy loss, while for AVD-Spatial, we adopted multi-label classification with binary cross-entropy to handle multiple spatial relations for a pair of objects. In the Proposal Ranking Module (PRM), we selected the top three boxes for Semantic Abstraction and the top ten for AVD-Spatial based on OPM. In our study, we set the threshold for DETIC proposal scores at 0.02, and for GroundingDINO, the bounding box threshold was 0.15, with the text threshold set at 0.10 to ensure optimal performance in the experiments.
\subsection{VLM Baselines}
\noindent
\textbf{LLaVA.}
We selected LLaVA \cite{liu2024visual} (Large Language and Vision Assistant) as a representative Vision-Language Model (VLM) due to its strong performance in zero-shot spatial reasoning, particularly for binary predicate prediction, as noted by SpatialVLM~\cite{Chen2024SpatialVLMEV}. To evaluate this, we prompted LLaVA with binary YES/NO questions such as, {\tt Is the {object$_1$} to the right of {object$_2$}? Answer with yes or no.} The F1 score and accuracy from these experiments are shown in Table \ref{table:MLP_AVD}, where LLaVA’s performance was found to be close to random chance compared to SRM experiments.
Additionally, we asked LLaVA to provide normalized bounding box coordinates for objects within captions, using prompts like, {\tt Give me the bounding box coordinates for the object.} We then calculated IoU accuracy and mean IoU, with results displayed in Table~\ref{table:final_results}. LLaVA’s generated bounding boxes tended to be overly large, resulting in lower IoU scores.


\setlength{\tabcolsep}{0.5\tabcolsep}
\begin{table}
\begin{center}
\begin{tabular}{@{}l*{7}{c}@{}}
\hline\noalign{\smallskip}

\multirow{2}{*}{Features} & \multicolumn{3}{c}{Top1 $\% \uparrow$ }  &  \multicolumn{3}{c}{Top2 $\% \uparrow$ } \\
        \cmidrule{2-7}
        &   Visual &  Synonyms  &  Class &    Visual &  Synonyms  &  Class   \\
\noalign{\smallskip}
\hline
\noalign{\smallskip}

 2D-Geom &  28.44 & 28.71 & 26.12  & 50.08 & 45.93 &  50.09\\
2D-Geom+Lng  & 36.54 & 35.59 & 41.26  & 60.14 & 60.14 & 63.47\\
\cmidrule{1-7}

3D-Geom  & 70.33 & 69.62 & 71.40  & 84.31 & 83.88 & 86.39\\
3D-Geom+Lng &  \textbf{76.16} &  \textbf{75.83} &  \textbf{78.01} &    \textbf{89.72} &  \textbf{89.76} &  \textbf{91.93}\\


\hline
\end{tabular}
\caption{The SRM results on the Semantic Abstraction dataset demonstrate effective classification of object pair relations. "2D-Geom" represents 2D geometric features, "3D-Geom" represents 3D geometric features, and "Lng" refers to language features. }
\label{table:cls_detic}
\end{center}
\end{table}

\begin{table*}
\begin{center}
\begin{tabular}{@{}l*{10}{c}@{}}
\hline\noalign{\smallskip}

 \multirow{2}{*}{Features} & \multicolumn{6}{c}{F1 $\% \uparrow$ }  &  \multirow{2}{*}{Micro F1 $\% \uparrow$ } & \multirow{2}{*}{Macro F1 $\% \uparrow$ } & \multirow{2}{*}{ACC $\% \uparrow$ } \\
        \cmidrule{2-7}
        &Behind&Front&Above &Below&Left&Right\\
\noalign{\smallskip}
\hline
\noalign{\smallskip}

2D-Geom & 14.60 & 15.93 & 32.97 & 42.77 & 90.37 & 90.86 & 74.05 & 47.91  &  91.77 \\
2D-Geom+Lng  & 27.40 & 29.23 & 67.40 & 69.68 & 91.10 & 100.00 & 82.93 & 64.14 & 94.14\\
\cmidrule{1-10}
3D-Geom  & 68.81 & 70.51 & 83.74  & 83.36 & 93.73 & 93.68 & 88.44 & 82.31 & 95.69\\
3D-Geom+Lng &  \textbf{71.34} &  \textbf{72.22} &  \textbf{86.86} & \textbf{88.02} &  \textbf{100.00} &  \textbf{100.00} & \textbf{93.62} &  \textbf{86.41} &  \textbf{97.63}\\
\cmidrule{1-10}

LLaVA-1.5-Vicuna-7B  & 35.99 & 40.25 & 66.16 & 58.35 & 57.12 & 65.62 &42.12& 53.92 & 42.12\\
LLaVA-1.5-Vicuna-13B  &3.91 & 2.62 & 6.11 & 2.19 & 34.52 & 24.64 & 11.63& 12.33 &  11.63\\


\hline
\end{tabular}
\caption{The results for SRM on AVD-Spatial, compared with LLaVA, demonstrate how the designed 3D geometric features can help classify the relationships between pairs of objects. }
\label{table:MLP_AVD}
\end{center}
\end{table*}
\setlength{\tabcolsep}{0.5\tabcolsep}

\commentOUT{JK:
\begin{table}
\begin{center}
\begin{tabular}{@{}l*{4}{c}@{}}
\hline\noalign{\smallskip}
%
\multirow{1}{*}{Supervision}& \multirow{1}{*}{Approach} & \multicolumn{1}{c}{Top1 $\% \uparrow$ }  &  \multicolumn{1}{c}{Top2 $\% \uparrow$ }&  \multicolumn{1}{c}{Top3 $\% \uparrow$ }\\
\noalign{\smallskip}
\hline
\noalign{\smallskip}
\multirow{1}{*}{--}&Language & 30.62 & 40.03 & 50.41 \\
\cmidrule{1-5}
\multirow{2}{*}{2D} & Geometric & 40.42 & 55.11& 66.04 \\
&Geometric + Language & 42.35 & 56.27& 66.81 \\
\cmidrule{1-5}
&
\multirow{2}{*}{2.5D} & Geometric & 43.24 &57.15 &  \textbf{68.19}\\
&Geometric + Language & \textbf{45.00} & {57.75}& 67.15 \\
\cmidrule{1-5}

\multirow{2}{*}{3D} & Geometric &  38.19 & 52.94 & 63.03 \\
&Geometric + Language & 44.62 &\textbf{57.76} & 68.07 \\
\hline
\end{tabular}
\caption{
In this table, we have presented the results of the SRM on real-world data. The results demonstrate how effectively we can classify the relation between pairs of objects in the SpatialSense dataset.}
\label{table:SRM_SS}
\end{center}
\end{table}
}

\commentOUT{
\setlength{\tabcolsep}{1.4\tabcolsep}
\begin{table*}
\begin{center}
\begin{tabular}{@{}l*{12}{c}@{}}
\hline\noalign{\smallskip}
%
\multirow{2}{*}{Approach} & \multirow{2}{*}{Language Model} &  \multicolumn{3}{c}{ACC $\% \uparrow$}  &  \multicolumn{3}{c}{Mean IOU $\uparrow$} \\
        \cmidrule{3-6}
\cmidrule{6-8}
        &&   Visual &  Synonyms  &  Class &   Visual &  Synonyms  &  Class  \\
\noalign{\smallskip}
\hline
\noalign{\smallskip}
\multirow{4}{*}{DETIC}  & -- & 24.97 & 25.31 & 22.26 & 0.23 &0.24 &0.20  \\
&Fast\_Text & 35.83 & 35.11 & 30.43 &0.33 & 0.33 & 0.28\\
&Bert+cls  & 37.92 & 38.44 & 30.95  &0.35 & 0.36 & 0.29\\
&Bert+avg  & {38.43} & {38.51} &{32.83} &{0.36} & {0.36} & {0.30}\\
&Target Object & \textbf{45.76} & \textbf{45.46} &\textbf{36.27} &\textbf{0.42} & \textbf{0.41} & 
\textbf{0.33}\\
\hline
\end{tabular}
\caption{This table demonstrates the results of the OPM module on Semantic Abstraction data, showcasing the effectiveness of our baseline. The results are shown for three different testing subsets proposed by the dataset. We have reported both the accuracy and the mean IOU to understand the effectiveness of the approach.}
\label{table:detic_as_grounding_SS}
\end{center}
\end{table*}
}

\setlength{\tabcolsep}{1.4\tabcolsep}
\begin{table}

\begin{center}
\begin{tabular}{@{}l*{7}{c}@{}}
\hline\noalign{\smallskip}
\multirow{2}{*}{Approach} & \multicolumn{3}{c}{ACC $\% \uparrow$}  &  \multicolumn{3}{c}{Mean IOU $\uparrow$} \\
        \cmidrule{2-7}
        & Visual &  Synonyms  &  Class &   Visual &  Synonyms  &  Class  \\
\noalign{\smallskip}
\hline
\noalign{\smallskip}

 OPM(DETIC) &   53.67 & 53.86& 46.49 & 0.49& 0.49 & 0.42 \\

 OPM(DETIC)+SRM+PRM &  \textbf{55.14}  & \textbf{55.46}  &  \textbf{47.53} & \textbf{0.51}& \textbf{0.51} &    \textbf{0.43} \\
\hline
 \multirow{1}{*}{Semantic Abstraction} & -- & --& -- & 0.19& 0.23 & 0.20 \\
\hline

\end{tabular}
\caption{We have demonstrated the performance of the entire pipeline across the three main test sets of the Semantic Abstraction. }
\label{table:final_results}
\end{center}
\end{table}

\setlength{\tabcolsep}{1.0\tabcolsep}
\begin{table}

\begin{center}
\begin{tabular}{@{}l*{7}{c}@{}}
\hline\noalign{\smallskip}
\multirow{1}{*}{Approach} &{ACC $\% \uparrow$}  & {Mean IOU $\uparrow$} \\
\noalign{\smallskip}
\hline
\noalign{\smallskip}

 OPM(DETIC) & 60.93 & 55.68 \\
 OPM(DETIC)+SRM+PRM &  \textbf{63.24} & \textbf{58.12}\\
\hline
 OPM(G-DINO) & 50.92 & 48.72  \\
 OPM(G-DINO)+SRM+PRM &  54.89 & 52.31 \\
\hline
 \multirow{1}{*}{LLaVA-1.5-Vicuna-7B}  & 8 & 4.33 \\
 \multirow{1}{*}{LLaVA-1.5-Vicuna-13B}  & 13 & 4.68 \\
\hline
\multirow{1}{*}{G-DINO}  & 42.45 & 40.72 \\

\hline
\end{tabular}
\caption{We have demonstrated the performance of the entire pipeline, along with LLaVA and Grounding-DINO, on AVD-Spatial.}
\label{table:final_results_AVD}
\vspace{-20pt}
\end{center}
\end{table}

\commentOUT{
\setlength{\tabcolsep}{1\tabcolsep}
\begin{table}
\begin{center}
\begin{tabular}{@{}l*{5}{c}@{}}
\hline\noalign{\smallskip}
%
\multirow{1}{*}{Approach} & \multirow{1}{*}{Classes}& \multirow{1}{*}{Language Model} &  \multicolumn{1}{c}{ACC $\% \uparrow$}  &  \multicolumn{1}{c}{Mean IOU $\uparrow$} \\
\noalign{\smallskip}
\hline
\noalign{\smallskip}
\multirow{2}{*}{Detic} &  Target Object & --& \textbf{45.50} & \textbf{0.42}  \\
& Lvis  & Bert+avg  & 35.06 & 0.33 \\
%
%
\hline
\end{tabular}
\caption{This table demonstrates the results of the OPM module on real-world data, Which demonstrates the effectiveness of our baseline.}
\label{table:detic_on_SS}
\end{center}
\end{table}
}

\begin{figure}[t]
  \centering
    \includegraphics[width=\linewidth]{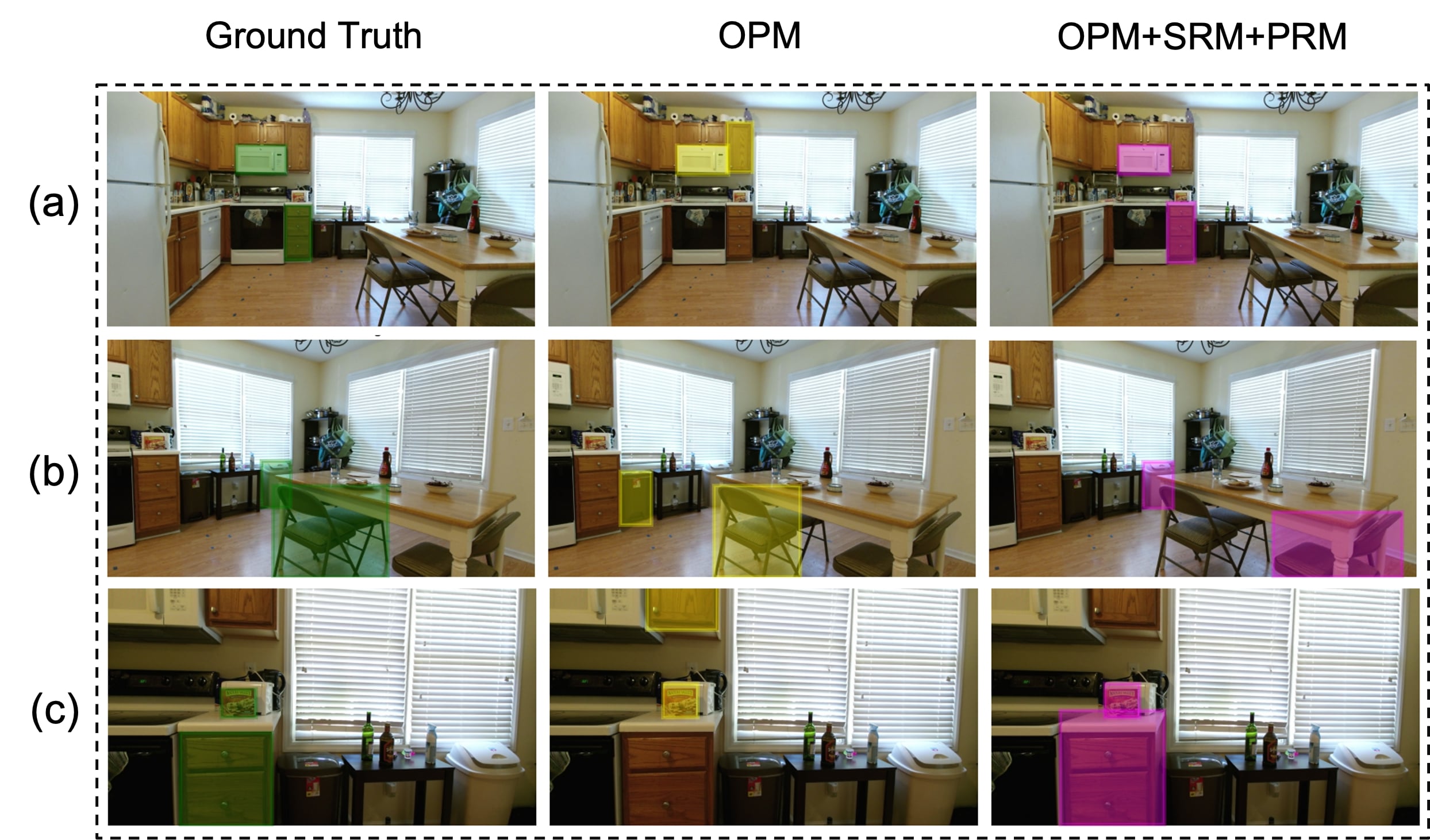}

    \caption{(a) The cabinet is below the microwave oven. (b) The trashcan is behind the chair. (c) The Nature Valley Sweet and Salty Nut Almond is above the cabinet. These results demonstrate how the SRM and PRM helped select a better pair as the target and reference objects.}
   \label{fig:AVD_visualizations}
   \vspace{-15pt}
\end{figure}

\begin{figure}[t]
  \centering
    \includegraphics[width=\linewidth]{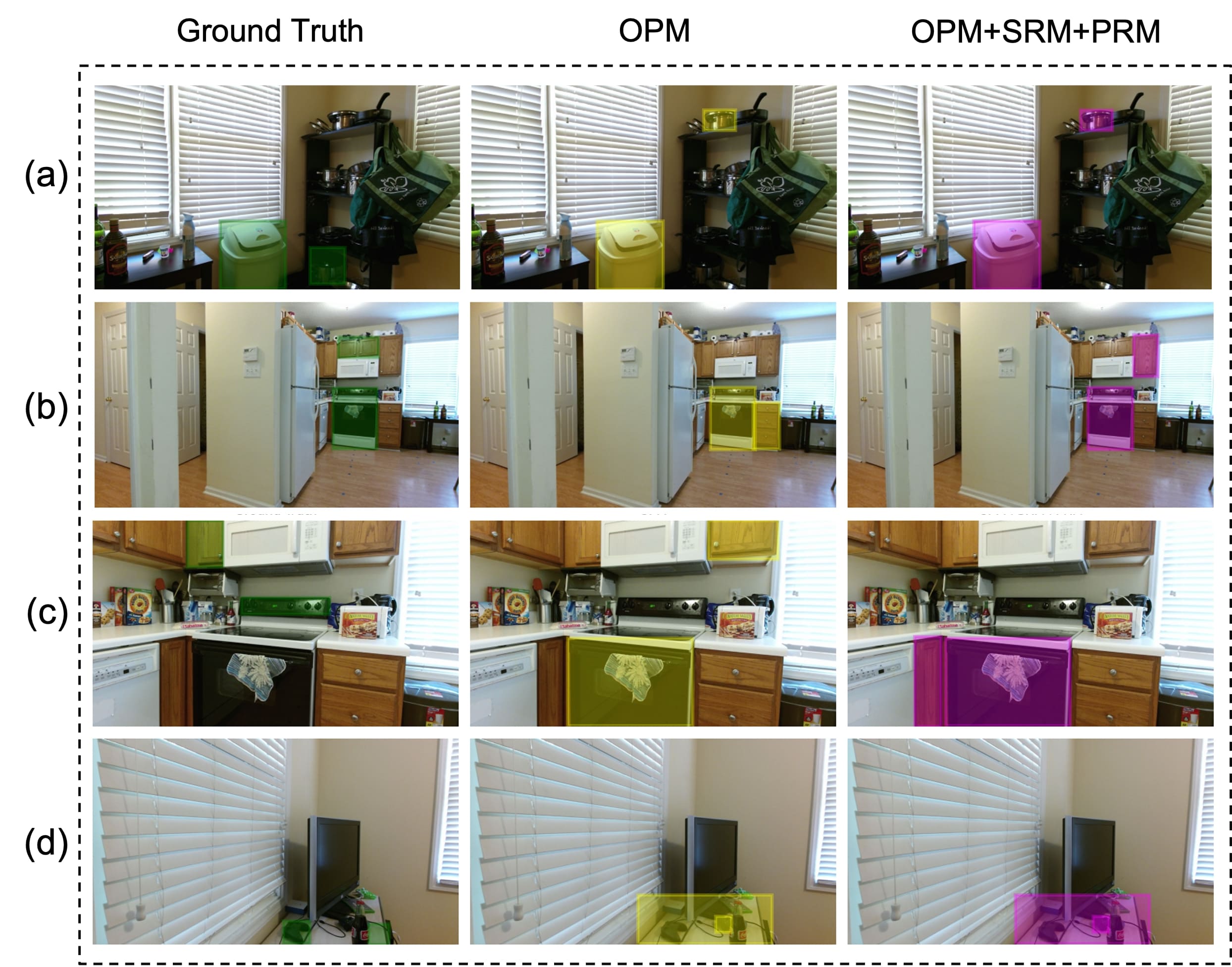}

    \caption{(a) The pot is right of the trashcan. (b) the cabinet is above the oven. (c) the cabinet is left of the oven. (d) the desk is right of the control. The labeling pipeline has two main issues: multiple correct answers, as seen in (a), (b), and (c), and noisy labels, and boxes. Despite this, based on qualitative evaluation, the proposed modular approach often performs more accurately when multiple correct answers are possible like (b) and (c).}
   \label{fig:labeling_errors}
\vspace{-15pt}
\end{figure}

\noindent
\textbf{GroundingDINO.} GroundingDINO is an advanced open-set object detector that integrates language-guided queries to detect arbitrary objects specified by category names or referring expressions. We evaluated Grounding DINO on our triplets, treating them as referring expressions, and reported the results in Table \ref{table:final_results_AVD}. The results demonstrate the superior performance of GroundingDINO compared to LLaVA, with an improvement of over 25\%. Grounding DINO was also used as the object detector in the OPM module, with the results included in the same table. 

\subsection{Probabilistic Spatial Relation Detection}
\label{sec:ModularspatialRelation}

\noindent
\textbf{SRM.} For the Spatial Relation Module (SRM), we assess performance based on accuracy for Semantic Abstraction and the F1 score for AVD-Spatial shown in Table \ref{table:cls_detic}, and for AVD-Spatial in Table \ref{table:MLP_AVD}. Our findings demonstrate a clear advantage of 3D over 2D geometric features. We further examined the role of language features in our classification module, showing their potential to boost performance. The results confirm SRM's superior performance over LLaVA. The experiments on the Semantic Abstraction dataset also emphasize the robustness of our MLP model, even when faced with visual or textual gaps. This resilience stems from the carefully selected geometric features and word embeddings, enabling the model to handle both visual and linguistic domain gaps effectively.\\
\noindent
\textbf{OPM+SRM+PRM.} To evaluate the proposed probabilistic structured approach, we have reported the accuracy of Intersection over Union (IoU) with threshold 0.5 compared to the ground-truth bounding box and mean IoU. The performance on the Semantic Abstraction dataset in Table \ref{table:final_results}, we have also demonstrated the OPM  performance to show the effectiveness of the SRM and PRM. The results show the robustness and effectiveness of the proposed ranking modules, outperforming the Semantic abstraction baseline presented in \cite{ha2022semantic}. This baseline presented the results for mean IoU in 3D voxel representations, replicating their results for 2D IOU is non-trivial. However, given use of ground-truth depth, and the considerable gap, our model performs better. Table \ref{table:final_results_AVD} demonstrates results on AVD-Spatial, showing superior performance over LLaVA and Grounding-DINO. By incorporating structured approach, our method is explainable, allowing us to diagnose module-specific failures. 
In contrast, LLaVA is more resilient due to its integrated design, though less interpretable. Figure \ref{fig:AVD_visualizations} visualized the results and highlighted how SRM and PRM improve object identification compared to relying solely on OPM. Figure \ref{fig:labeling_errors}, shows some labeling flaws of the automated approach, our structured method still provides more satisfactory results compared to using only the OPM module.

\section{CONCLUSIONS}
Our paper introduces a probabilistic approach for spatial relations recognition. Through extensive experiments on real and synthetic datasets, we demonstrate the effectiveness of integrating 3D geometric cues and open set object detectors, showing superior performance compared to zero-shot spatial reasoning of LLaVa model. The dataset component can be used for generating training data for fine-tuning VLM's that is a subject of future work. The presented approach can handle various domain shifts and diverse environments, offering a robust solution applicable to robotics settings.

\bibliographystyle{IEEEtran}
\bibliography{mainSR}

\begin{thebibliography}{10}
\providecommand{\url}[1]{#1}
\csname url@rmstyle\endcsname
\providecommand{\newblock}{\relax}
\providecommand{\bibinfo}[2]{#2}
\providecommand\BIBentrySTDinterwordspacing{\spaceskip=0pt\relax}
\providecommand\BIBentryALTinterwordstretchfactor{4}
\providecommand\BIBentryALTinterwordspacing{\spaceskip=\fontdimen2\font plus
\BIBentryALTinterwordstretchfactor\fontdimen3\font minus \fontdimen4\font\relax}
\providecommand\BIBforeignlanguage[2]{{%
\expandafter\ifx\csname l@#1\endcsname\relax
\typeout{** WARNING: IEEEtran.bst: No hyphenation pattern has been}%
\typeout{** loaded for the language `#1'. Using the pattern for}%
\typeout{** the default language instead.}%
\else
\language=\csname l@#1\endcsname
\fi
#2}}

\bibitem{DBLP:journals/corr/AmmiratoPPKB17}
\BIBentryALTinterwordspacing
P.~Ammirato, P.~Poirson, E.~Park, J.~Kosecka, and A.~C. Berg, ``A dataset for developing and benchmarking active vision,'' \emph{CoRR}, vol. abs/1702.08272, 2017. [Online]. Available: \url{http://arxiv.org/abs/1702.08272}
\BIBentrySTDinterwordspacing

\bibitem{ha2022semantic}
H.~Ha and S.~Song, ``Semantic abstraction: Open-world 3d scene understanding from 2d vision-language models,'' 2022.

\bibitem{Chen2024SpatialVLMEV}
B.~Chen, Z.~Xu, S.~Kirmani, B.~Ichter, D.~Driess, P.~Florence, D.~Sadigh, L.~Guibas, and F.~Xia, ``Spatialvlm: Endowing vision-language models with spatial reasoning capabilities,'' \emph{ArXiv}, vol. abs/2401.12168, 2024.

\bibitem{mo2022towards}
Y.~Mo, H.~Zhang, and T.~Kong, ``Towards open-world interactive disambiguation for robotic grasping,'' in \emph{CoRL 2022 Workshop on Learning, Perception, and Abstraction for Long-Horizon Planning}, 2022.

\bibitem{Yuan2024RoboPointAV}
\BIBentryALTinterwordspacing
W.~Yuan, J.~Duan, V.~Blukis, W.~Pumacay, R.~Krishna, A.~Murali, A.~Mousavian, and D.~Fox, ``Robopoint: A vision-language model for spatial affordance prediction for robotics,'' \emph{ArXiv}, vol. abs/2406.10721, 2024. [Online]. Available: \url{https://api.semanticscholar.org/CorpusID:270559475}
\BIBentrySTDinterwordspacing

\bibitem{Jiang2023HierarchicalRA}
B.~Jiang and C.~J. Taylor, ``Hierarchical relationships: A new perspective to enhance scene graph generation,'' 2023.

\bibitem{nejatishahidin2023graph}
N.~Nejatishahidin, W.~Hutchcroft, M.~Narayana, I.~Boyadzhiev, Y.~Li, N.~Khosravan, J.~Ko{\v{s}}eck{\'a}, and S.~B. Kang, ``Graph-covis: Gnn-based multi-view panorama global pose estimation,'' in \emph{Proceedings of the IEEE/CVF Conference on Computer Vision and Pattern Recognition}, 2023, pp. 6458--6467.

\bibitem{Lu2019ViLBERTPT}
J.~Lu, D.~Batra, D.~Parikh, and S.~Lee, ``Vilbert: Pretraining task-agnostic visiolinguistic representations for vision-and-language tasks,'' in \emph{Neural Information Processing Systems}, 2019.

\bibitem{DBLP:journals/corr/abs-2112-03857}
L.~H. Li, P.~Zhang, H.~Zhang, J.~Yang, C.~Li, Y.~Zhong, L.~Wang, L.~Yuan, L.~Zhang, J.~Hwang, K.~Chang, and J.~Gao, ``Grounded language-image pre-training,'' \emph{CoRR}, vol. abs/2112.03857, 2021.

\bibitem{Tan2019LXMERTLC}
H.~H. Tan and M.~Bansal, ``Lxmert: Learning cross-modality encoder representations from transformers,'' in \emph{Conference on Empirical Methods in Natural Language Processing}, 2019.

\bibitem{Li2021AlignBF}
J.~Li, R.~R. Selvaraju, A.~D. Gotmare, S.~R. Joty, C.~Xiong, and S.~C.~H. Hoi, ``Align before fuse: Vision and language representation learning with momentum distillation,'' in \emph{Neural Information Processing Systems}, 2021.

\bibitem{liu2024improved}
H.~Liu, C.~Li, Y.~Li, and Y.~J. Lee, ``Improved baselines with visual instruction tuning,'' in \emph{Proceedings of the IEEE/CVF Conference on Computer Vision and Pattern Recognition}, 2024, pp. 26\,296--26\,306.

\bibitem{liu2024visual}
H.~Liu, C.~Li, Q.~Wu, and Y.~J. Lee, ``Visual instruction tuning,'' \emph{Advances in neural information processing systems}, vol.~36, 2024.

\bibitem{team2023gemini}
G.~Team, R.~Anil, S.~Borgeaud, Y.~Wu, J.-B. Alayrac, J.~Yu, R.~Soricut, J.~Schalkwyk, A.~M. Dai, A.~Hauth, \emph{et~al.}, ``Gemini: a family of highly capable multimodal models,'' \emph{arXiv preprint arXiv:2312.11805}, 2023.

\bibitem{anil2023palm}
R.~Anil, A.~M. Dai, O.~Firat, M.~Johnson, D.~Lepikhin, A.~Passos, S.~Shakeri, E.~Taropa, P.~Bailey, Z.~Chen, \emph{et~al.}, ``Palm 2 technical report,'' \emph{arXiv preprint arXiv:2305.10403}, 2023.

\bibitem{radford2021learning}
A.~Radford, J.~W. Kim, C.~Hallacy, A.~Ramesh, G.~Goh, S.~Agarwal, G.~Sastry, A.~Askell, P.~Mishkin, J.~Clark, \emph{et~al.}, ``Learning transferable visual models from natural language supervision,'' in \emph{International conference on machine learning}.\hskip 1em plus 0.5em minus 0.4em\relax PMLR, 2021, pp. 8748--8763.

\bibitem{alayrac2022flamingo}
J.-B. Alayrac, J.~Donahue, P.~Luc, A.~Miech, I.~Barr, Y.~Hasson, K.~Lenc, A.~Mensch, K.~Millican, M.~Reynolds, \emph{et~al.}, ``Flamingo: a visual language model for few-shot learning,'' \emph{Advances in neural information processing systems}, vol.~35, pp. 23\,716--23\,736, 2022.

\bibitem{kamath-etal-2023-whats}
\BIBentryALTinterwordspacing
A.~Kamath, J.~Hessel, and K.-W. Chang, ``What{'}s {``}up{''} with vision-language models? investigating their struggle with spatial reasoning,'' in \emph{Proceedings of the 2023 Conference on Empirical Methods in Natural Language Processing}, H.~Bouamor, J.~Pino, and K.~Bali, Eds.\hskip 1em plus 0.5em minus 0.4em\relax Singapore: Association for Computational Linguistics, Dec. 2023, pp. 9161--9175. [Online]. Available: \url{https://aclanthology.org/2023.emnlp-main.568}
\BIBentrySTDinterwordspacing

\bibitem{Rajabi2023TowardsGV}
N.~Rajabi and J.~Kosecka, ``Towards grounded visual spatial reasoning in multi-modal vision language models,'' \emph{ArXiv}, vol. abs/2308.09778, 2023.

\bibitem{Liu2022VisualSR}
F.~Liu, G.~E.~T. Emerson, and N.~Collier, ``Visual spatial reasoning,'' \emph{Transactions of the Association for Computational Linguistics}, vol.~11, pp. 635--651, 2022.

\bibitem{Chen_2024_CVPR}
B.~Chen, Z.~Xu, S.~Kirmani, B.~Ichter, D.~Sadigh, L.~Guibas, and F.~Xia, ``Spatialvlm: Endowing vision-language models with spatial reasoning capabilities,'' in \emph{Proceedings of the IEEE/CVF Conference on Computer Vision and Pattern Recognition (CVPR)}, June 2024, pp. 14\,455--14\,465.

\bibitem{Driess2023PaLMEAE}
\BIBentryALTinterwordspacing
D.~Driess, F.~Xia, M.~S.~M. Sajjadi, C.~Lynch, A.~Chowdhery, B.~Ichter, A.~Wahid, J.~Tompson, Q.~H. Vuong, T.~Yu, W.~Huang, Y.~Chebotar, P.~Sermanet, D.~Duckworth, S.~Levine, V.~Vanhoucke, K.~Hausman, M.~Toussaint, K.~Greff, A.~Zeng, I.~Mordatch, and P.~R. Florence, ``Palm-e: An embodied multimodal language model,'' in \emph{International Conference on Machine Learning}, 2023. [Online]. Available: \url{https://api.semanticscholar.org/CorpusID:257364842}
\BIBentrySTDinterwordspacing

\bibitem{Johnson2016CLEVRAD}
\BIBentryALTinterwordspacing
J.~Johnson, B.~Hariharan, L.~van~der Maaten, L.~Fei-Fei, C.~L. Zitnick, and R.~B. Girshick, ``Clevr: A diagnostic dataset for compositional language and elementary visual reasoning,'' \emph{2017 IEEE Conference on Computer Vision and Pattern Recognition (CVPR)}, pp. 1988--1997, 2016. [Online]. Available: \url{https://api.semanticscholar.org/CorpusID:15458100}
\BIBentrySTDinterwordspacing

\bibitem{Yang2019SpatialSenseAA}
K.~Yang, O.~Russakovsky, and J.~Deng, ``Spatialsense: An adversarially crowdsourced benchmark for spatial relation recognition,'' \emph{2019 IEEE/CVF International Conference on Computer Vision (ICCV)}, pp. 2051--2060, 2019.

\bibitem{Goyal2020Rel3DAM}
A.~Goyal, K.~Yang, D.~Yang, and J.~Deng, ``Rel3d: A minimally contrastive benchmark for grounding spatial relations in 3d,'' \emph{ArXiv}, vol. abs/2012.01634, 2020.

\bibitem{Kolve2017AI2THORAI}
\BIBentryALTinterwordspacing
E.~Kolve, R.~Mottaghi, W.~Han, E.~VanderBilt, L.~Weihs, A.~Herrasti, M.~Deitke, K.~Ehsani, D.~Gordon, Y.~Zhu, A.~Kembhavi, A.~K. Gupta, and A.~Farhadi, ``Ai2-thor: An interactive 3d environment for visual ai,'' \emph{ArXiv}, vol. abs/1712.05474, 2017. [Online]. Available: \url{https://api.semanticscholar.org/CorpusID:28328610}
\BIBentrySTDinterwordspacing

\bibitem{zhou2022detecting}
X.~Zhou, R.~Girdhar, A.~Joulin, P.~Kr{\"a}henb{\"u}hl, and I.~Misra, ``Detecting twenty-thousand classes using image-level supervision,'' in \emph{ECCV}, 2022.

\bibitem{Liu2023GroundingDM}
\BIBentryALTinterwordspacing
S.~Liu, Z.~Zeng, T.~Ren, F.~Li, H.~Zhang, J.~Yang, C.~yue Li, J.~Yang, H.~Su, J.-J. Zhu, and L.~Zhang, ``Grounding dino: Marrying dino with grounded pre-training for open-set object detection,'' \emph{ArXiv}, vol. abs/2303.05499, 2023. [Online]. Available: \url{https://api.semanticscholar.org/CorpusID:257427307}
\BIBentrySTDinterwordspacing

\bibitem{clip}
A.~Radford, J.~W. Kim, C.~Hallacy, A.~Ramesh, G.~Goh, S.~Agarwal, G.~Sastry, A.~Askell, P.~Mishkin, J.~Clark, \emph{et~al.}, ``Learning transferable visual models from natural language supervision,'' in \emph{International conference on machine learning}.\hskip 1em plus 0.5em minus 0.4em\relax PMLR, 2021, pp. 8748--8763.

\bibitem{Subramanian2022ReCLIPAS}
S.~Subramanian, W.~Merrill, T.~Darrell, M.~Gardner, S.~Singh, and A.~Rohrbach, ``Reclip: A strong zero-shot baseline for referring expression comprehension,'' in \emph{Annual Meeting of the Association for Computational Linguistics}, 2022.

\bibitem{gpv}
T.~Gupta, A.~Kamath, A.~Kembhavi, and D.~Hoiem, ``Towards general purpose vision systems: An end-to-end task-agnostic vision-language architecture,'' in \emph{Proceedings of the IEEE/CVF Conference on Computer Vision and Pattern Recognition (CVPR)}, June 2022, pp. 16\,399--16\,409.

\bibitem{mdetr}
A.~Kamath, M.~Singh, Y.~LeCun, G.~Synnaeve, I.~Misra, and N.~Carion, ``Mdetr-modulated detection for end-to-end multi-modal understanding,'' in \emph{Proceedings of the IEEE/CVF International Conference on Computer Vision}, 2021, pp. 1780--1790.

\bibitem{DBLP:journals/corr/KrishnaZGJHKCKL16}
\BIBentryALTinterwordspacing
R.~Krishna, Y.~Zhu, O.~Groth, J.~Johnson, K.~Hata, J.~Kravitz, S.~Chen, Y.~Kalantidis, L.~Li, D.~A. Shamma, M.~S. Bernstein, and L.~Fei{-}Fei, ``Visual genome: Connecting language and vision using crowdsourced dense image annotations,'' \emph{CoRR}, vol. abs/1602.07332, 2016. [Online]. Available: \url{http://arxiv.org/abs/1602.07332}
\BIBentrySTDinterwordspacing

\bibitem{Kirillov2023SegmentA}
A.~Kirillov, E.~Mintun, N.~Ravi, H.~Mao, C.~Rolland, L.~Gustafson, T.~Xiao, S.~Whitehead, A.~C. Berg, W.-Y. Lo, P.~Doll{\'a}r, and R.~B. Girshick, ``Segment anything,'' \emph{2023 IEEE/CVF International Conference on Computer Vision (ICCV)}, pp. 3992--4003, 2023.

\bibitem{Li2023LabelingIS}
\BIBentryALTinterwordspacing
Y.~Li, N.~Rajabi, S.~Shrestha, M.~A. Reza, and J.~Kosecka, ``Labeling indoor scenes with fusion of out-of-the-box perception models,'' \emph{2024 IEEE/CVF Winter Conference on Applications of Computer Vision Workshops (WACVW)}, pp. 570--579, 2023. [Online]. Available: \url{https://api.semanticscholar.org/CorpusID:269203025}
\BIBentrySTDinterwordspacing

\bibitem{Cheng2021PerPixelCI}
\BIBentryALTinterwordspacing
B.~Cheng, A.~G. Schwing, and A.~Kirillov, ``Per-pixel classification is not all you need for semantic segmentation,'' in \emph{Neural Information Processing Systems}, 2021. [Online]. Available: \url{https://api.semanticscholar.org/CorpusID:235829267}
\BIBentrySTDinterwordspacing

\end{thebibliography}
\end{document}